\documentclass[10pt,twocolumn,letterpaper]{article}

\usepackage[pagenumbers]{cvpr} 

\usepackage{colortbl}
\definecolor{cvprblue}{rgb}{0.21,0.49,0.74}
\definecolor{lightblue}{rgb}{0.9,0.94,1}
\definecolor{mycolor_blue}{HTML}{E7EFFA}
\definecolor{mycolor_green}{HTML}{E6F8E0}
\definecolor{mycolor_gray}{HTML}{ECECEC}
\definecolor{pearDark}{HTML}{2980B9}
\usepackage[pagebackref,breaklinks,colorlinks,allcolors=cvprblue]{hyperref}
\usepackage{multirow}
\usepackage[accsupp]{axessibility}
\usepackage{algorithm}
\usepackage{algpseudocode} %
\usepackage{setspace}
\usepackage{colortbl}
\usepackage{graphicx}
\usepackage{animate}
\usepackage{caption}
\def\paperID{8393} 
\def\confName{CVPR}
\def\confYear{2025}
\newcommand{\wyk}[1]{\textcolor{black}{#1}}

\newcommand{\camera}[1]{\textcolor{black}{#1}}
\title{VideoDirector: Precise Video Editing via Text-to-Video Models}

\author{
	Yukun Wang$^{1}$ \quad Longguang Wang$^{1}$ \quad Zhiyuan Ma$^{2}$ \quad Qibin Hu$^{1}$ \quad Kai Xu$^{3}$ \quad Yulan Guo$^{1*}$ \\
	\vspace{-0.8em} \\
	{ $^1$Shenzhen Campus of Sun Yat-sen University, Sun Yat-sen University \quad $^2$Tsinghua University}\\  
    {$^3$National University of Defense Technology}\\
	{\tt\small wangyk59@mail2.sysu.edu.cn, wanglg9@mail.sysu.edu.cn, guoyulan@sysu.edu.cn}\\
	{Project webpage:~\href{https://yukun66.github.io/VideoDirector/}{https://VideoDirector.com}}
}

\begin{document}
\twocolumn[{%
    \maketitle
    \vspace{-3em} 
    \begin{center}
        \centering
        \includegraphics[width=\textwidth]{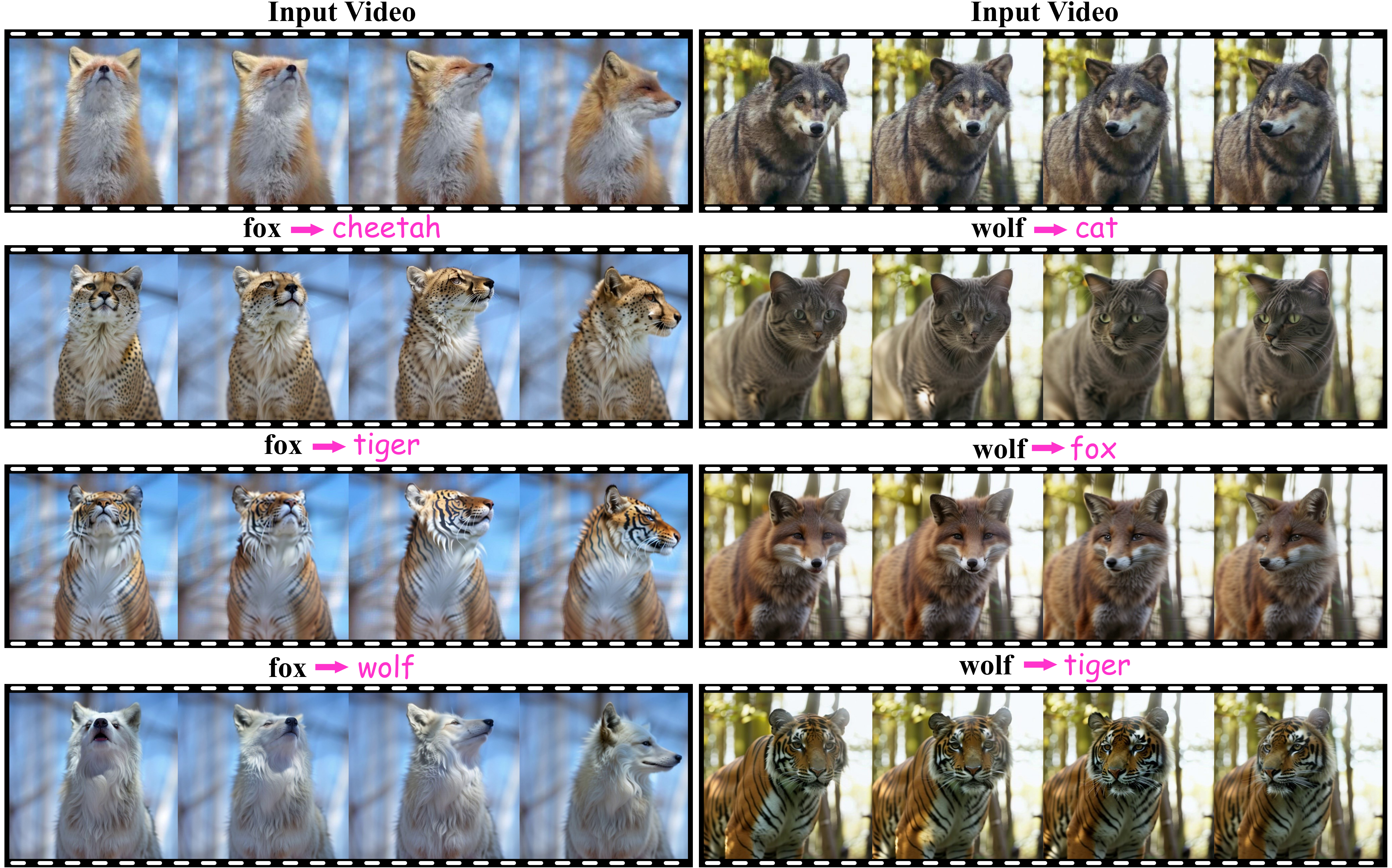} 
        \captionof{figure}{
        \textbf{\textit{Edited results}}. Our method enables precise content editing of an input video based on a text prompt, while preserving unedited content. By directly leveraging the text-to-video (T2V) generation model~\cite{guo2023animatediff} for editing,  the edited results exhibit high fidelity, real-world motion smoothness, and enhanced realism.}
        \label{fig: teaser}
    \end{center}
    \vspace{0.2em} 
}]
\renewcommand{\thefootnote}{\fnsymbol{footnote}}
\footnotetext[1]{Corresponding author.}
\begin{abstract}
Despite the typical inversion-then-editing paradigm using text-to-image (T2I) models has demonstrated promising results, directly extending it to text-to-video (T2V) models still suffers severe artifacts
such as color flickering and content distortion. Consequently, current video editing methods primarily rely on T2I models, which inherently lack temporal-coherence generative ability, often resulting in inferior editing results. 
In this paper, we attribute the failure of the typical editing paradigm to: 
\wyk{1) \textbf{Tightly Spatial-temporal Coupling.} 
The vanilla pivotal-based inversion strategy struggles to disentangle spatial-temporal information in the video diffusion model; 
2) \textbf{Complicated Spatial-temporal Layout.} 
 The vanilla cross-attention control is deficient in preserving the unedited content.} 
To address these limitations, 
\wyk{we propose a spatial-temporal decoupled guidance~(STDG) and multi-frame null-text optimization strategy to provide pivotal temporal cues for more precise pivotal inversion. Furthermore, we introduce a self-attention control strategy to maintain higher fidelity for precise partial content editing.}
Experimental results demonstrate that our method (termed VideoDirector) effectively harnesses the powerful temporal generation capabilities of T2V models, producing edited videos with state-of-the-art performance in accuracy, motion smoothness, realism, and fidelity to unedited content. 

\end{abstract}    
\begin{figure*}[t] 
  \centering
  \begin{subfigure}[t]{0.48\linewidth} 
    \centering
    \includegraphics[width=\linewidth]{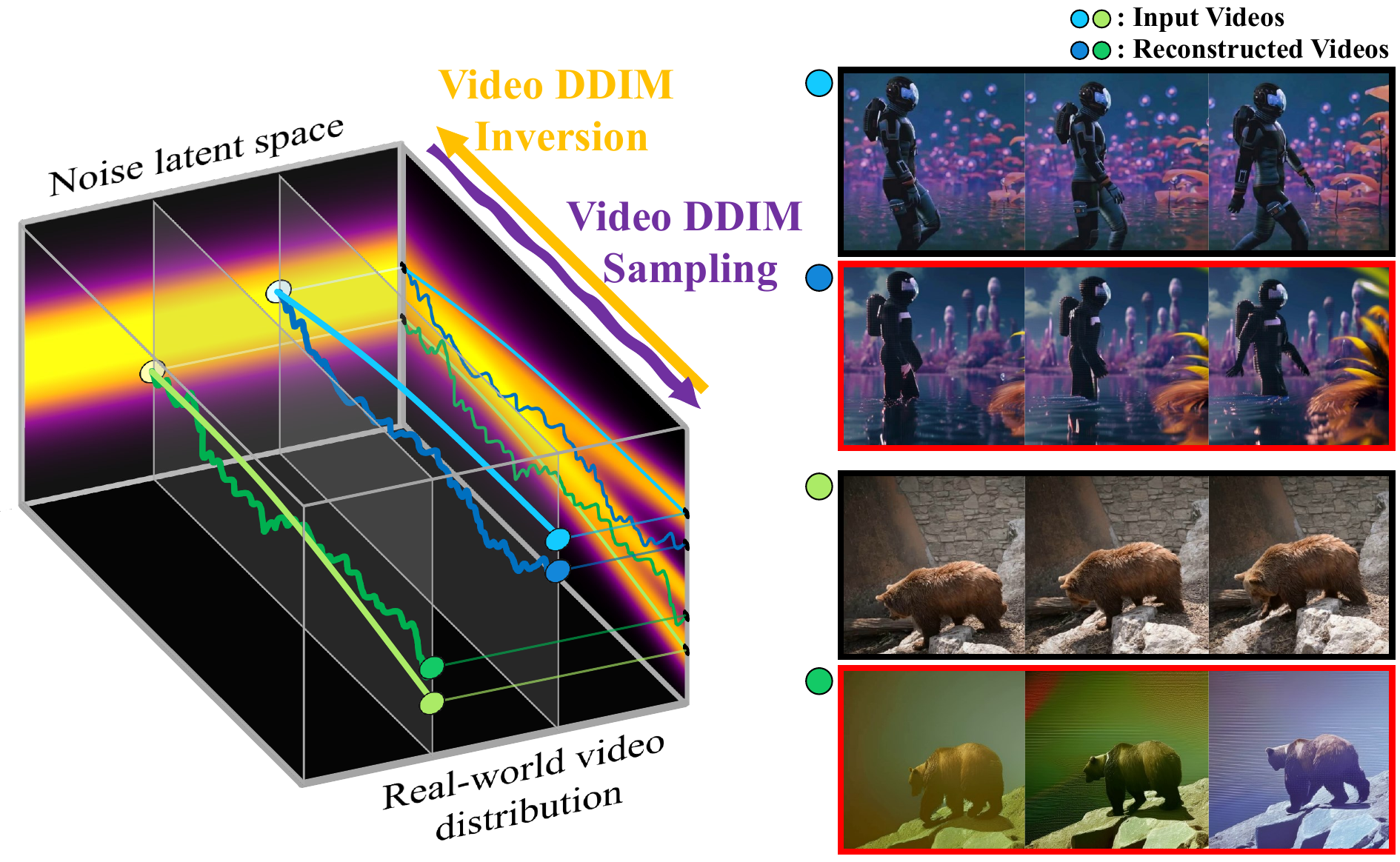}
    \caption{The prompt-to-prompt~\cite{hertz2022prompt} and null-text optimization~\cite{mokady2022null} are integrated directly into the T2V generation model~\cite{guo2023animatediff} to reconstruct the input videos. The results present challenges for the typical editing paradigm~\cite{hertz2022prompt, mokady2022null} in accurately reconstructing the original videos.}
    \label{fig:motivation_a}
  \end{subfigure}
  \hfill
  \begin{subfigure}[t]{0.48\linewidth}
    \centering
    \includegraphics[width=\linewidth]{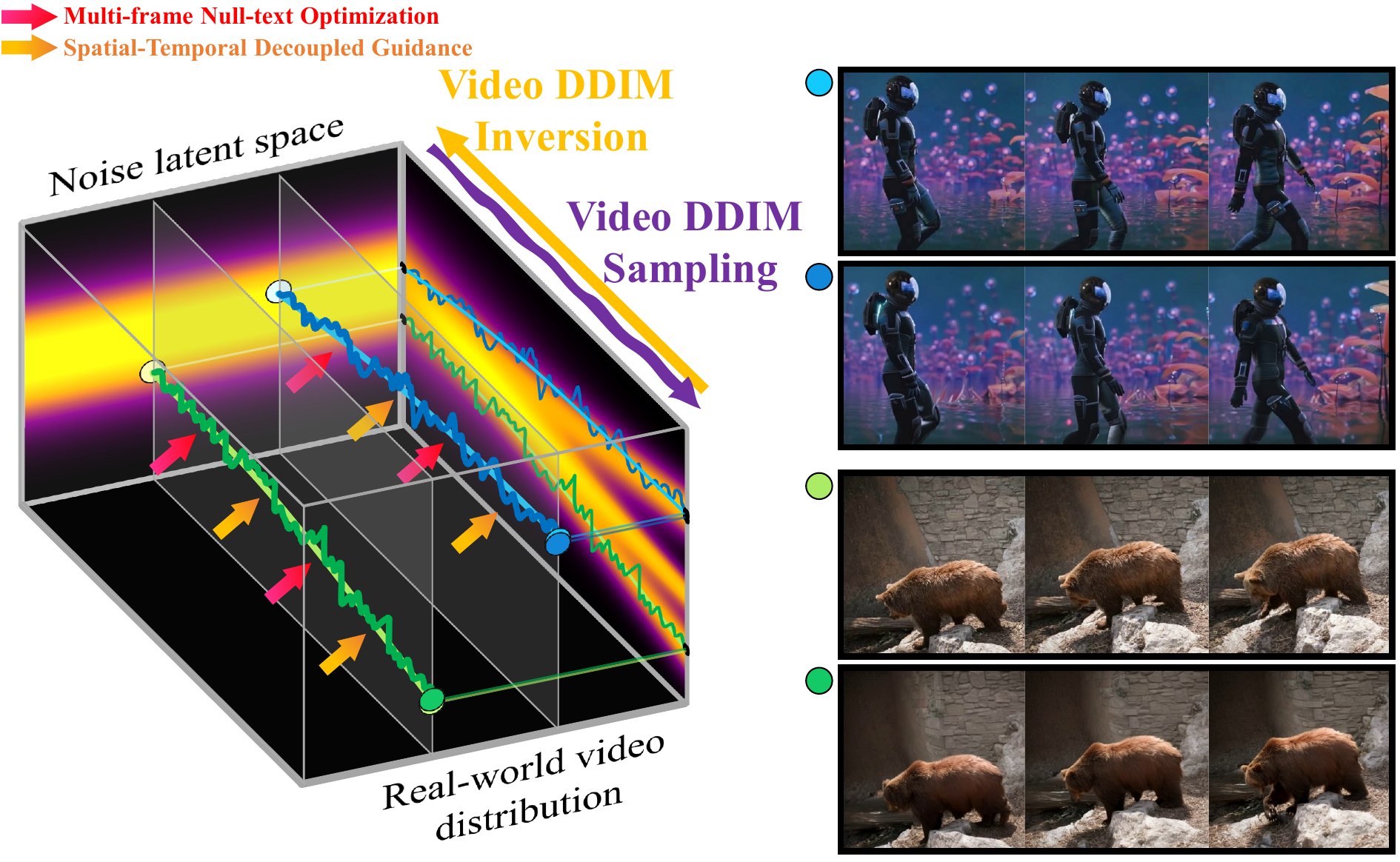}
    \caption{Our method achieves accurate reconstruction of input videos by incorporating multi-frame Null-text optimization and spatial-temporal decoupled guidance. 
    }
    \label{fig:motivation_b}
  \end{subfigure}
  \vspace{-0.5em}
  \caption{\textbf{\textit{Principle visualization of our approach.}} Comparison of diffusion pivotal inversion~\cite{mokady2022null} using a T2V generation model~\cite{guo2023animatediff} integrated with vanilla null-text optimization (a) and our proposed guidance (b). Our approach constrains the reverse diffusion trajectory during video generation to align with DDIM inversion, enabling precise reconstruction of the input video.}
  \label{fig: motivation}
  \vspace{-1em}
\end{figure*}

\section{Introduction}
\label{sec:intro}
With the advancement of diffusion models~\cite{ho2020denoising, song2020denoising, luo2022understanding}, recent years have witnessed significant progress of generative networks, particularly in text-to-image (T2I) generation~\cite{rombach2022high, ho2022classifier, saharia2022photorealistic} and text-to-video (T2V) generation communities~\cite{chen2024videocrafter2, guo2023animatediff, ma2024latte, blattmann2023align}. Motivated by their success, a series of image editing~\cite{hertz2022prompt, mokady2022null, wu2024turboedit, shi2024dragdiffusion, ruiz2023dreambooth, han2023svdiff} and video editing~\cite{geyer2023tokenflow, liu2024video, zhang2024camel, ling2024motionclone, cong2024flatten, kara2024rave} methods have been proposed to achieve visual content editing via text prompts, promoting a wide range of applications. 
Notably, instead of using T2V models, current video editing methods are still built upon T2I models by leveraging inter-frame features~\cite{geyer2023tokenflow, kara2024rave, li2024vidtome}, incorporating optical flows~\cite{cong2024flatten}, or training auxiliary temporal layers~\cite{liu2024video}. As a result, these methods still suffer inferior realism and temporal coherence due to the absence of temporal coherence in vanilla T2I models.
This raises a question: \emph{Can we edit a video directly using T2V models?} 

In the field of image editing, the typical ``\emph{inversion-then-editing}" paradigm mainly includes two steps: pivotal inversion and attention-controlled editing. \wyk{First, unbiased pivotal inversion is achieved by null-text optimization and classifier-free guidance~\cite{mokady2022null}. Then, content editing is performed using a cross-attention control strategy~\cite{hertz2022prompt}.}
Despite the success in T2I models, directly applying this
paradigm to T2V models often leads to significant deviations from the original input, such as severe color flickering and background variations in \cref{fig:motivation_a}. 
%
%

In this paper, we attribute these failures to: 1) Tightly spatial-temporal coupling. The entanglement of temporal and spatial (appearance) information in T2V models prevents vanilla pivotal inversion from compensating for the biases introduced by DDIM inversion.
2) Complicated spatial-temporal layout. The vanilla cross-attention control is insufficient to maintain the complex spatial-temporal layout of video content, resulting in low-fidelity editing results.


To address these issues, we first introduce an auxiliary spatial-temporal decoupled guidance (STDG) to provide additional temporal cues. Simultaneously, we extend shared null-text embeddings to a multi-frame strategy to accommodate temporal information. These components alleviate the bias from the DDIM inversion, enabling the diffusion backward trajectory to be accurately aligned with the initial trajectory, as shown in~\cref{fig:motivation_b}. In addition, we propose a self-attention control strategy to maintain complex spatial-temporal layout and enhance editing fidelity.

\camera{Our key insight is that \textit{accurate reconstruction is the basis of high-quality video editing}, where \textit{accurate reconstruction} (\cref{sec:pivotal inversion}) refers to precisely resampling the input video from noise using T2V models, and \textit{video editing} (\cref{sec: Attention Control}) can be formulated as precisely controlling the reconstruction trajectory to achieve the desired modifications. Previously, directly integrating DDIM inversion-based methods with T2V models cannot produce satisfactory results~\cref{fig:motivation_a}. Our approach demonstrates the feasibility of controlling the denoising trajectory for precise editing via T2V model.}

Overall, our contributions are summarized as:
\begin{itemize}
\item We introduce spatial-temporal decoupled guidance (STDG) and \wyk{multi-frame null-text optimization} to provide temporal cues \wyk{for pivotal inversion in T2V model.}
\item \wyk{We develop a self-attention control strategy to maintain the complex spatial-temporal layout and enhance fidelity. Our approach demonstrates the feasibility of controlling the denoising trajectory for precise editing.}
\item Extensive experiments demonstrate that our method effectively utilizes T2V models for video editing, significantly outperforming state-of-the-art methods in fidelity, motion smoothness and realism.
\end{itemize}
\section{Related Work}
\label{sec:relatedwork}



\noindent\textbf{Text-to-Image Editing }
Recent advances in T2I generation models have promoted the rapid development of text-guided image editing methods~\cite{rombach2022high, hertz2022prompt, mokady2022null, ma2024adapedit, ruiz2023dreambooth, shi2024dragdiffusion, wu2024turboedit}. 
Hertz \etal~\cite{hertz2022prompt} introduced Prompt-to-Prompt to edit images via DDIM inversion and manipulation of cross-attention maps. Specifically, techniques such as Word Swap, Phrase Addition, and Attention Re-weighting are performed to modify the attention maps based on text prompts. 
Since the DDIM inversion introduces biases by approximating noise latent,
Mokady \etal~\cite{mokady2022null} introduced a step-wise null-text embedding $\phi_t$ optimized after DDIM inversion for compensation.
This optimization refines the denoising trajectory by compensating for DDIM inversion biases, enhancing both reconstruction quality and editing precision.
Different from this pipeline, DreamBooth~\cite{ruiz2023dreambooth} fine-tuned a pre-trained T2I model~\cite{saharia2022photorealistic} to synthesize subjects in prompt-guided diverse scenes using reference images as additional conditions. 

\noindent\textbf{Text-to-Video Editing}
Numerous efforts have been made to extend T2I models directly to video editing~\cite{liu2024video, cong2024flatten, geyer2023tokenflow,kara2024rave, chai2023stablevideo, cohen2024slicedit}. Tune-A-Video~\cite{wu2023tune} developed a tailored spatio-temporal attention mechanism and an efficient one-shot tuning strategy to tune an input video. Video-P2P~\cite{liu2024video} transforms a T2I model to a video-customized Text-to-Set (T2S) model through fine-tuning to achieve semantic consistency across adjacent frames. 
TokenFlow~\cite{geyer2023tokenflow} explicitly propagates token features based on inter-frame correspondences using the T2I model without any additional training or fine-tuning. 
RAVE~\cite{kara2024rave} utilizes Controlnet and introduces random shuffling of latent grids to ensure temporal consistency.
Flatten~\cite{cong2024flatten} incorporates optical flow into the attention module of the T2I model to address inconsistency issues in text-to-video editing. Due to the lack of temporal generation capacity of T2I models, the aforementioned methods still suffer from results with inferior temporal coherence, realism, and motion smoothness. 

\section{Method}
\label{sec:method}
\begin{figure*}[t] 
  \centering
  \includegraphics[width=0.99\linewidth]{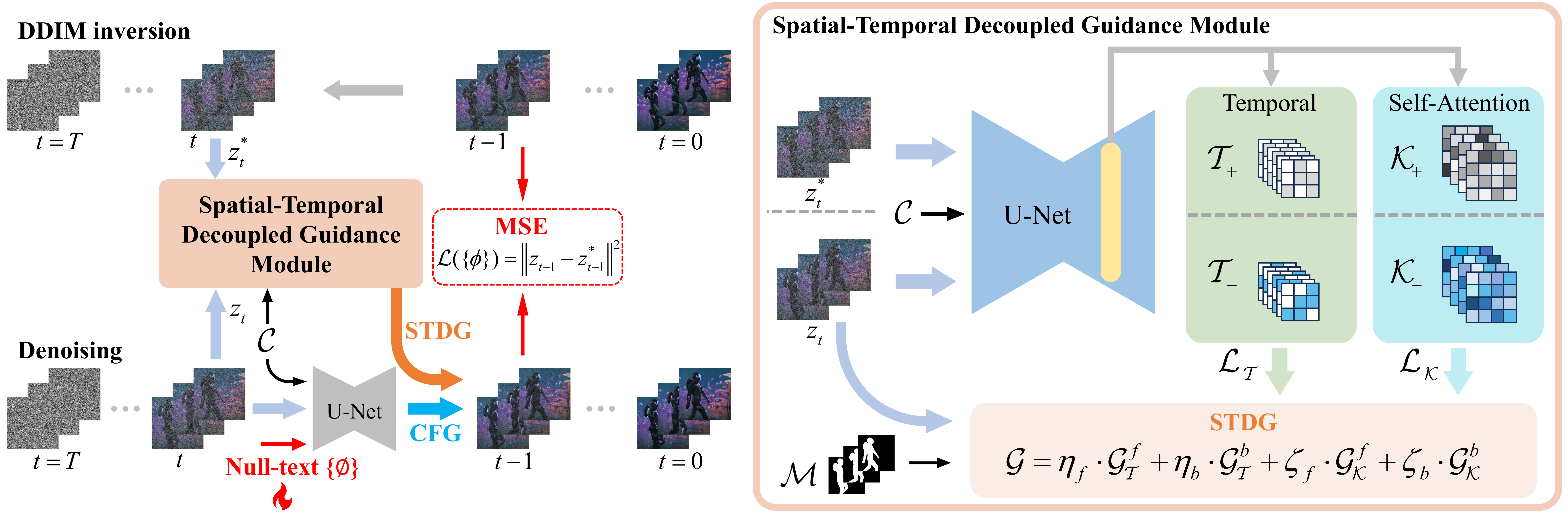} 
  \vspace{-0.5em}
  \caption{\textbf{\textit{Video pivotal inversion pipeline.}} Our pipeline comprises two key components: multi-frame null-text optimization and spatial-temporal decoupled guidance, which are integrated into the standard pivotal inversion pipeline. 
  }
  \label{fig:optimize_pipeline}
  \vspace{-1em}
\end{figure*}

\subsection{Preliminaries}
\paragraph{Latent Diffusion Models~(LDMs).}
In LDM~\cite{rombach2022high}, the forward process generates a noisy image latent $\boldsymbol z_t$ by combining the original image $\boldsymbol z_0$ with Gaussian noise $\epsilon_t$:
\vspace{-3pt}
\begin{equation}
\boldsymbol z_t = \sqrt{\alpha_t} \boldsymbol z_0 + \sqrt{1-\alpha_t}\epsilon_t, ~ where ~~ \epsilon_t \sim \mathcal N(\boldsymbol 0, \boldsymbol I),
\label{eq: LDM1}
\vspace{-3pt}
\end{equation}
where $\boldsymbol z_0$ is the image latent encoded by the VAE encoder $\mathcal E(\cdot)$.
During training, given the noisy latent $z_t$ and condition $c$ such as text, the diffusion model $\epsilon_{\theta}$ is encouraged to predict the noise $\epsilon_t$ at step $t$: 
\vspace{-3pt}
\begin{equation}
\mathcal L(\theta) = \mathbb{E}_{\mathcal E(x), \epsilon \sim \mathcal N(\boldsymbol 0, \boldsymbol I), t \sim \mathcal U(1,T)} [\Vert \epsilon_t - \epsilon_{\theta}(\boldsymbol z_t, c, t) \Vert_2^2].
\label{eq: LDM2}
\vspace{-3pt}
\end{equation}
During inference, given a condition $c$, the model iteratively
samples $\boldsymbol z_{t-1}$ from $\boldsymbol z_{t}$
using the diffusion model. 
Classifier-free guidance (CFG)~\cite{ho2022classifier}
are employed to guide the sampling trajectory: 
\vspace{-3pt}
\begin{equation}
\hat{\epsilon_{\theta}} = \epsilon_{\theta}(z_t, c, t)+\omega[\epsilon_{\theta}(z_t, c, t)-\epsilon_{\theta}(z_t, \phi, t)],
\label{eq: CFG}
\vspace{-3pt}
\end{equation}
where $\omega$ is guidance weight, and $\phi$ represents null-text or a negative prompt. 

\noindent\textbf{DDIM Sampling and Inversion.}
\label{sec: DDIM_sampling_and_inversion}
DDIM~\cite{song2020denoising} provides a more efficient sampling strategy with only tens of steps.
Given the latent $z_t$, the transition from $z_t$ to $z_{t-1}$ is derived using predicted noise $\epsilon_\theta(\boldsymbol z_t)$:
\begin{equation}
\scalebox{0.85}{$
\boldsymbol z_{t-1} = \sqrt{\alpha_{t-1}}
\left( \dfrac{\boldsymbol z_t-\sqrt{1-\alpha_t}\epsilon_\theta(\boldsymbol z_t)}{\sqrt{\alpha_t}}\right)
+ \sqrt{1-\alpha_{t-1}}\epsilon_\theta(\boldsymbol z_t).
$}
\label{eq: DDIM sampling}
\end{equation}
Then, we can derive a transformation that expresses $z_t$ in terms of $z_{t-1}$, and shift $(t)$ or $(t-1)$ to $(t+1)$ and $(t)$. This allows us to obtain the DDIM inversion:
\begin{equation}
\scalebox{0.85}{$
\boldsymbol z_{t+1} = \sqrt{\alpha_{t+1}}
\left( \dfrac{\boldsymbol z_t-\sqrt{1-\alpha_t}\epsilon_\theta(\boldsymbol z_t)}{\sqrt{\alpha_t}}\right) 
+ \sqrt{1-\alpha_{t+1}}\epsilon_\theta(\boldsymbol z_t).
$}
\label{eq: DDIM inversion}
\end{equation}
Since $\epsilon_\theta(\boldsymbol z_{t+1})$ cannot be obtained without $\boldsymbol z_{t+1}$, it is approximated by $\epsilon_\theta(\boldsymbol z_{t})$. This approximation limits the ability to fully recover the original content when performing denoising solely from the noisy latents of DDIM inversion.

\noindent\textbf{Diffusion Pivotal Inversion.}
As discussed above, the approximation during DDIM inversion introduces deviations, causing the trajectory of denoising latents to deviate from the ideal no-bias DDIM inversion. To address this, Mokady \etal~\cite{mokady2022null} introduced a step-wise null-text embedding $\phi_t$ optimized after DDIM inversion: 
\begin{equation}
\mathcal L(\phi_t) = \Vert z_{t-1}^* - z_{t-1} \Vert_2^2, 
\label{eq: null_text}
\end{equation}
where $z_t$ and $z_t^*$ represent the latents from denoising and DDIM inversion, respectively. This optimization refines the denoising trajectory by compensating for DDIM inversion biases, enhancing both reconstruction and editing quality.

\subsection{Pivotal Inversion for Video Reconstruction}
\label{sec:pivotal inversion}
Despite promising results in T2I images, directly applying pivotal inversion techniques~\cite{hertz2022prompt, mokady2022null} to T2V models still suffer severe deviation from the original trajectory, as illustrated in~\cref{fig:motivation_a}. 
We attribute this deviation to two reasons. \textit{First}, vanilla null-text embeddings share itself across all video frames and lack temporal modeling capability. \textit{Second}, vanilla classifier-free guidance is insufficient for distinguishing temporal cues from spatial ones, resulting in meaningless latents.
With an additional temporal dimension, fine-grained temporal awareness is required for precise manipulation of the latent in T2V models.
To this end, we propose multi-frame null-text embeddings and spatial-temporal decoupled guidance.

\noindent\textbf{Multi-Frame Null-Text Embeddings.\label{sec:multi_frame_null_text_optimization}}
To accommodate additional temporal information in the video, we introduce multi-frame null-text embeddings ($\{\boldsymbol{\phi}_t\}\in\mathbb{R}^{F\times l\times c}$), where $l$ and $c$ represent the sequence length and embedding dimension, as illustrated in Fig.~\ref{fig:optimize_pipeline}. Compared with vanilla null-text embeddings, multi-frame null-text embeddings produce notable gains in terms of both accuracy and realism, as demonstrated in \cref{sec: ablation}.

\noindent\textbf{Spatial-Temporal Decoupled Guidance.\label{sec:STDG}}
Diffusion pivotal inversion~\cite{mokady2022null} has demonstrated its effectiveness in  meaningful image editing. However, due to the absence of temporal awareness, the pivotal noise vectors in T2V models fail to provide sufficient temporal information during pivotal inversion, resulting in meaningless outputs.
Inspired by MotionClone~\cite{ling2024motionclone}, we leverage the temporal and self-attention features during video pivotal inversion to obtain spatial-temporal decoupled guidance.

Intuitively, temporal coherence in the original video can be maintained by minimizing the difference between the temporal attention maps contained in the pivotal inversion process~(\cref{fig:optimize_pipeline}):
\begin{equation}
\begin{aligned}
&\mathcal L_{\mathcal T} = \mathcal M^{f/b}_{\mathcal T} \cdot \mathcal M_{\mathcal T} \cdot \Vert{(\mathcal T_{+} - \mathcal T_{-})}\Vert_2^2, \\
&\mathcal G_{\mathcal T}^{f/b} = \dfrac{\partial ( {\mathcal L_{\mathcal T}}) }{\partial z_t},
\end{aligned}
\label{eq: L_temp}
\end{equation}
where $\mathcal T_{+}$,
$\mathcal T_{-}\in\mathbb{R}^{(H*W*C)\times F\times F}$ denote the temporal attention maps
of DDIM inversion and denoising latents. 
Mask $\mathcal M_{\mathcal T}$ select the top $K$ values within the last dimension of these attention maps $\mathcal T$.
$\mathcal M^{f/b}_{\mathcal T}$ represents the foreground or background mask generated by the SAM2 model~\cite{ravi2024sam2}, reshaped to match the dimensions of the temporal attention weights. The gradient with respect to the denoised latent $z_t$ is then used as the temporal-aware guidance.

Similarly, spatial (appearance) consistency can be derived by minimizing the difference between the self-attention keys during pivotal inversion (\cref{fig:optimize_pipeline}): 
\begin{equation}
\begin{aligned}
&\mathcal L_{\mathcal K} = \mathcal M^{f/b}_{\mathcal K} \cdot \Vert{ (\mathcal K_{+} - \mathcal K_{-})}\Vert_2^2,\\
&\mathcal G_{\mathcal K}^{f/b} = \dfrac{\partial ( {\mathcal L_{\mathcal K}}) }{\partial z_t},
\end{aligned}
\label{eq: L_spatial}
\end{equation}
where $\mathcal K_{+}$,
$\mathcal K_{-}\in\mathbb{R}^{F\times (H*W)\times C}$ represent the self-attention keys of DDIM inversion and denoising latents, respectively. $\mathcal M^{f/b}_{\mathcal K}$ denotes the SAM2 mask reshaped to match the dimensions of the keys.
Overall, the spatial-temporal decoupled guidance can be obtained as:
\begin{equation}
\mathcal G = \eta_f \cdot \mathcal G_{\mathcal T}^{f} + \eta_b \cdot \mathcal G_{\mathcal T}^{b} + \zeta_f \cdot \mathcal G_{\mathcal K}^{f} + \zeta_b \cdot \mathcal G_{\mathcal K}^{b},
\label{eq: total_score}
\end{equation}
where $\eta_f$, $\eta_b$, $\zeta_f$, and $\zeta_b$ are the coefficients of the foreground and background decoupled guidance. Our proposed guidance explicitly disentangles the appearance and temporal information to provide more precise guidance for optimization while maintaining meaningful results. Finally, the STDG guides video generation trajectory together with CFG for more precise pivotal inversion and editing: 
\begin{equation}
\hat{\epsilon_{\theta}} = \epsilon_{\theta}(z_t, c, t)+\omega[\epsilon_{\theta}(z_t, c, t)-\epsilon_{\theta}(z_t, \phi, t)] + \mathcal{G},
\label{eq: guide}
\end{equation}
where $\omega$ is CFG guidance weight, and $\phi$ represents null-text or a negative prompt.
\subsection{Attention Control for Video Editing} 
\label{sec: Attention Control}
Based on effective video pivotal inversion, directly applying the cross-attention control strategy in T2I methods~\cite{hertz2022prompt, mokady2022null} still struggles to provide sufficient control for video editing \wyk{as the complicated relationship between spatial-temporal tokens. As a result, edited videos still suffer from inconsistent motion and deficiency in preserving unedited content, producing results with low fidelity to the original video.}
To address this issue, we introduce an attention control strategy tailored for video editing from the perspectives of both self-attention and cross-attention.
\begin{figure}[t] 
  \centering
  \includegraphics[width=0.99\linewidth]{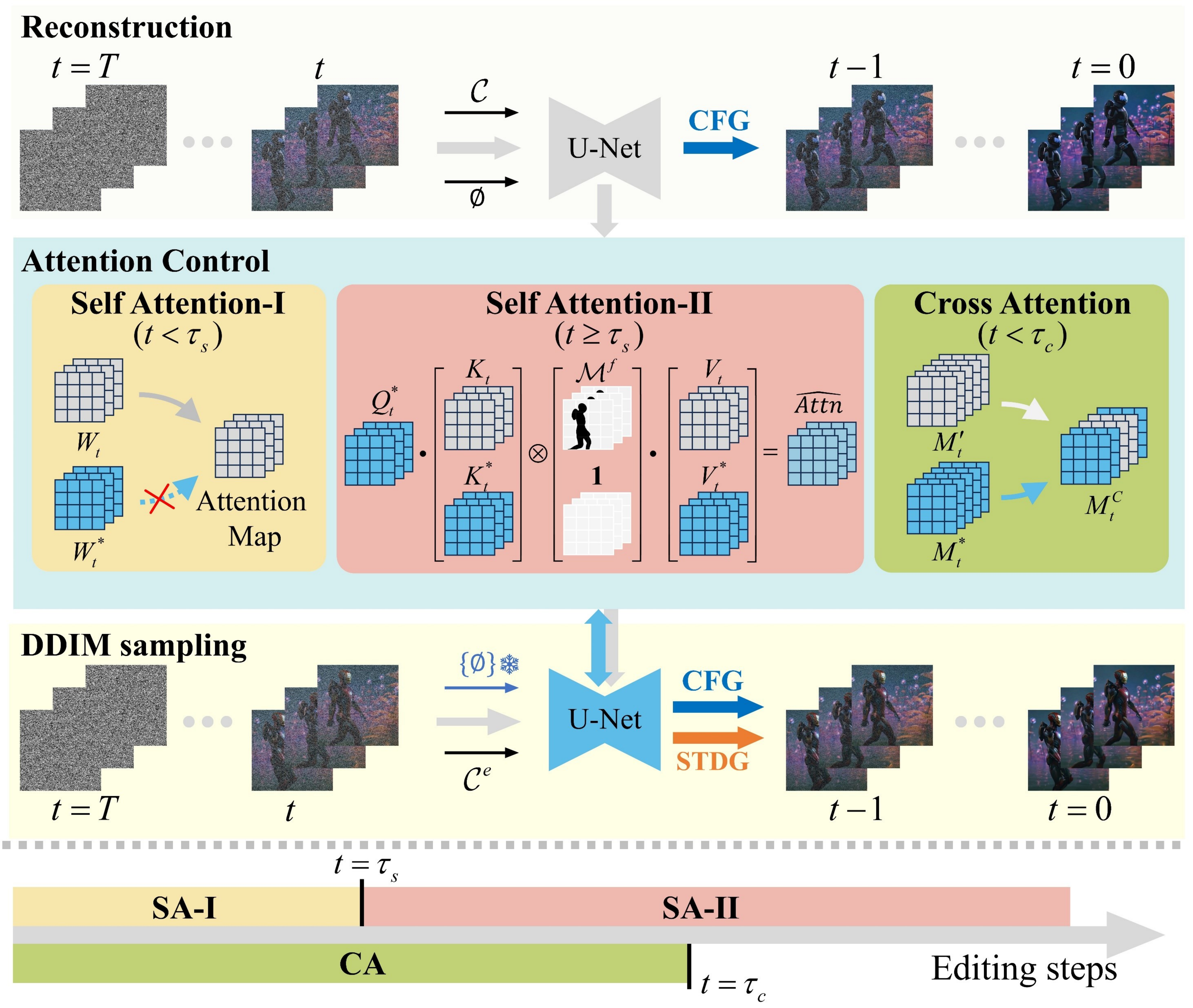} 
  \caption{\textbf{\textit{Our video editing pipeline.}} The \wyk{SA-\uppercase\expandafter{\romannumeral1} and SA-\uppercase\expandafter{\romannumeral2} maintain the complicated spatial-temporal layout and enhance fidelity, }while the cross-attention control introduces editing guidance based on the editing prompts.}
  \label{fig:editing_pipeline}
  \vspace{-1.5em}
\end{figure}

\noindent\textbf{Self-Attention Control.}
As illustrated in~\cref{fig:editing_pipeline}, we first introduce a self-attention-\uppercase\expandafter{\romannumeral1}~(\textbf{SA-\uppercase\expandafter{\romannumeral1}}) control strategy to \wyk{initialize the spatial-temporal layout aligning with the input video.}
At the beginning of editing, we replace the self-attention maps in the editing path with those from the reconstruction path during the first $\tau_s$ steps. 
\wyk{To further maintain the complicated spatial-temporal layout and enhance fidelity during editing,}
in self-attention-\uppercase\expandafter{\romannumeral2}~(\textbf{SA-\uppercase\expandafter{\romannumeral2}}), the self-attention keys~$K_t$, $K_t^*$
and values~$V_t$, $V_t^* \in\mathbb{R}^{F\times (H*W)\times C}$ from the reconstruction and editing processes are concatenated to obtain $\hat{K}_t = [K_t^* \mid K_t]$ 
and $\hat{V}_t = [V_t^* \mid V_t]\in\mathbb{R}^{F\times (2*H*W)\times C}$. Next, attention maps are calculated using the queries in the editing path and $\hat{K}_t$.
To prevent the incorporation of original content in the regions to be edited,  attention masks $\mathcal{M}^{f}$ derived from the SAM2 model~\cite{ravi2024sam2} is employed on the attention maps \wyk{to derive the mutual attentions}:
\begin{equation}
\widehat{Attn} = 
\begin{cases}
    W_t \cdot V_t^*, \quad \text{if } t < \tau_s, \vspace{0.2cm} \\
    S\left( \displaystyle \frac{Q_t^* \cdot \hat{K}_t^{\top}}{\sqrt{d}} \otimes \left[ \mathbf{1} \mid \mathcal{M}^{f} \right] \right) \cdot \hat{V}_t, &\text{otherwise.}\\
\end{cases}
\label{eq: self_attn_control}
\end{equation}
Here, $S$ represents the softmax operation. Finally, the resultant self-attention map is adopted to aggregate the values $\hat{V}_t$. The frame-wise attention mask $\mathcal{M}^{f}$ decouples edited and unedited content in the input video, enabling more precise and fine-grained editing. 
\wyk{This mutual attention module integrates keys and values from both paths in the editing pipeline, enhancing the preservation of complex spatial-temporal layouts and improving the harmony between edited and unedited contents.
Consequently, our self-attention control module enhances the fidelity of both motion and unedited content.}

\noindent\textbf{Cross-Attention Control.}
In addition to the self-attention control strategy, a cross-attention control strategy is employed during the first $\tau_c$ iterations to introduce information from the editing prompt into the latent. Specifically, for words common to both the editing prompt and the original prompt (\emph{i.e.}, ``walks with ... alien
plants that glow''), we replace the cross-attention maps in the editing path $M_t^*$ with those from the reconstruction path $M_t$. Meanwhile, the attention maps for novel words (\emph{i.e.}, ``Iron Man''), which are unique in the editing prompt, are retained in the editing path to introduce editing guidance. Finally, the cross-attention map $M_t^C$ is defined as follows:
\begin{equation}
M_t^C = 
\begin{cases}
    \boldsymbol{C} \cdot [\boldsymbol{\gamma} \cdot (M_t^*) + (\boldsymbol{1} - \boldsymbol{\gamma}) \cdot (M_t')], & \text{if } t < \tau_c,\\
    M_t^*, \quad \text{otherwise.}
\end{cases}
\label{eq: cross_attn_control}
\end{equation}
Here, $M_t'$ is mapped from $M_t$ for varying editing prompt lengths.  $\boldsymbol{\gamma}$ represents the binary vector used to combine the attention maps, while $\boldsymbol{C}$ denotes the re-weighting coefficient corresponding to each word in the editing prompts. 


\section{Experiments}
\label{sec: experiments}
\paragraph{Datasets and Baselines.}
\begin{figure*}[t]
    \centering
    \includegraphics[width=\textwidth]{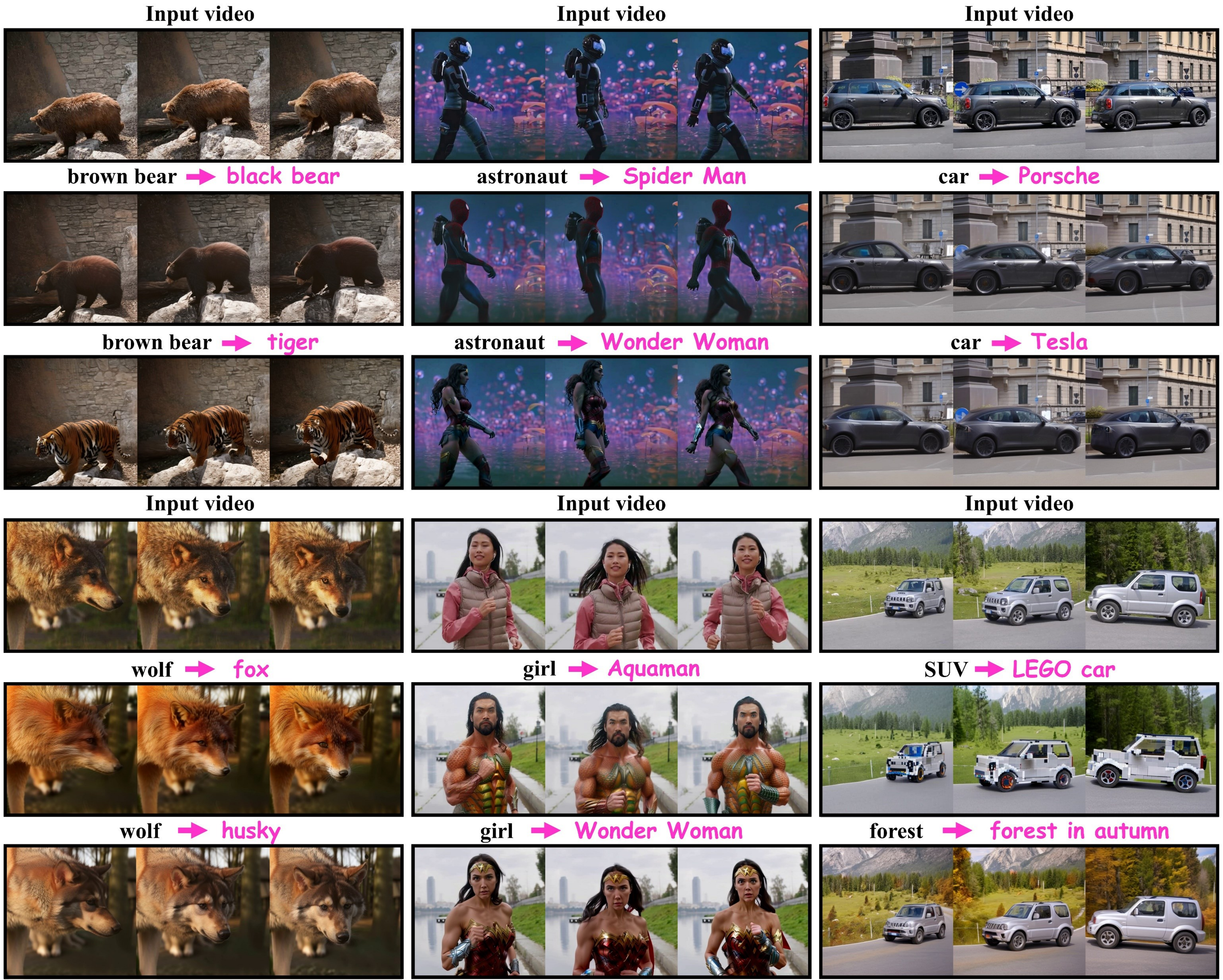} 
    \caption{\textbf{\textit{Edited results}}. The edited videos demonstrate our method's effectiveness in terms of accuracy, fidelity, motion smoothness, and realism. Moreover, the edited videos illustrate superior harmony, seamlessly integrating the edited content into the original unedited environment and context.}
    \label{fig: main_experiment}
    \vspace{-1.5em}
\end{figure*}
\begin{figure*}[t]
    \centering
    \includegraphics[width=\textwidth]{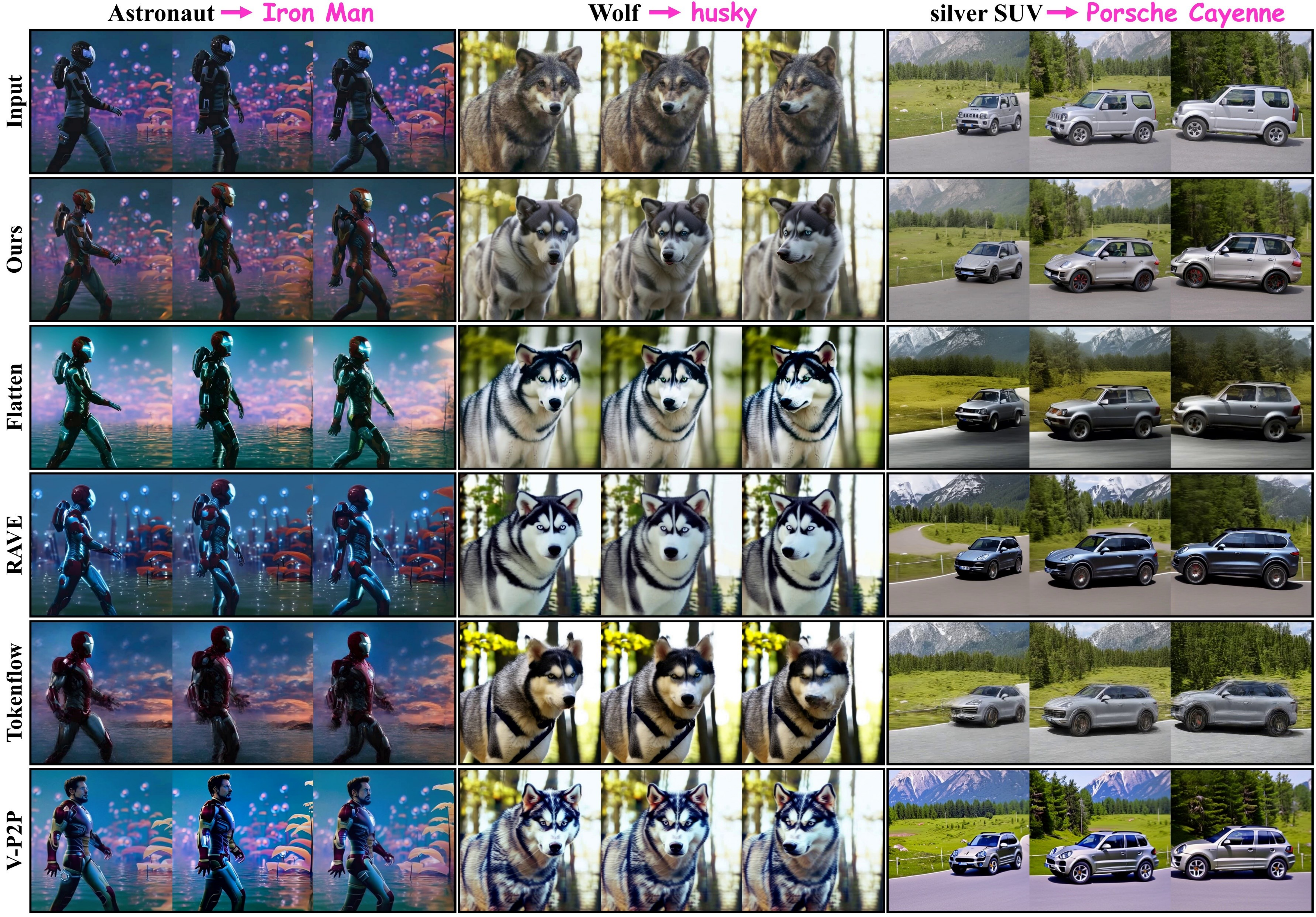} 
    \vspace{-1.5em}
    \caption{\textbf{\textit{Qualitative comparison}}. Our method achieves superior motion smoothness and realism compared to other approaches. We encourage readers to watch our video demo in supplementary material to observe the dynamic performance.}
    \vspace{-1em}
    \label{fig: Comparison}
\end{figure*}
We collected $75$ text-video editing pairs with a resolution of $512\times512$, including the videos sourced from the DAVIS dataset~\cite{perazzi2016benchmark}, MotionClone, Tokenflow~\cite{ling2024motionclone, geyer2023tokenflow}, and online platforms. The prompts are derived from ChatGPT or contributed by the authors. The videos utilized in our experiments cover diverse categories, including people, animals, and manual objects. 
%
We compare our approach with four state-of-the-art video editing methods based on T2I models, including Video-P2P~\cite{liu2024video}, 
RAVE~\cite{kara2024rave}, 
Flatten~\cite{cong2024flatten}, 
and Tokenflow~\cite{geyer2023tokenflow}. 
Video-P2P requires training a video-customized text-to-set~(T2S) model, which increases the editing time. 
RAVE enforces temporal consistency by randomly shuffling latent grids, while Flatten uses optical flow to improve temporal consistency. 

\noindent\textbf{Implementation Details.}
We implemented our method using AnimateDiff~\cite{guo2023animatediff} as the base T2V model. The number of video frames is fixed to $16$ due to the high memory consumption of AnimateDiff. Our method requires $8.5$ minutes for pivotal tuning and $1$ minute for video editing on a single A100 GPU. 
The cross-attention threshold ($\tau_c$ in Eq.~\ref{eq: self_attn_control}) was set to $0.8$, while the self-attention threshold ($\tau_s$ in Eq.~\ref{eq: cross_attn_control}) was manually tuned conditioned on the input video within the range of $[0.2, 0.5]$. For foreground editing, the coefficient $\eta_f$ was set to $0.5$, and $\eta_b$ was set between $0.2$ and $0.8$ in~\cref{eq: total_score}, $\zeta_f$ was set to $0$, and $\zeta_b$ to $0.5$. When editing the background, these values were swapped.

\subsection{Evaluation}
\begin{table}[bh]
    \centering
    \small 
    \scalebox{0.975}{ 
    \begin{tabular}{lccccc}
        \toprule
        \textbf{Methods} & MS~\textuparrow & PS~\textuparrow & m.P~\textuparrow & m.L~\textdownarrow & US~\textdownarrow \\
        \midrule
        Flatten~\cite{cong2024flatten} & 96.08\% & 21.24 & 14.70 & 0.329 & 3.11\\
        RAVE~\cite{kara2024rave} & 95.98\% & \cellcolor{mycolor_green}{\underline{21.61}} & 17.49 & 0.344 & \cellcolor{mycolor_green}{\underline{2.89}}\\
        Tokenflow~\cite{geyer2023tokenflow} & \cellcolor{mycolor_green}{\underline{96.69\%}} & 21.44 & \cellcolor{mycolor_green}{\underline{17.94}} & \cellcolor{mycolor_green}{\underline{0.313}} & 4.22\\
        V-P2P~\cite{liu2024video} & 94.46\% & 21.22 &17.66  & 0.340 & 3.78\\
        
         \textbf{Ours} & \cellcolor{lightblue}{\textbf{97.68\%}} & \cellcolor{lightblue}{\textbf{21.64}} & \cellcolor{lightblue}{\textbf{21.37}} & \cellcolor{lightblue}{\textbf{0.270}} & \cellcolor{lightblue}{\textbf{1}}\\
        \bottomrule
    \end{tabular}
    }
    \vspace{-1em}
    \caption{Comparison results across various metrics. We highlight the best value in \colorbox{lightblue}{\textbf{blue}}, and the second-best value in \colorbox{mycolor_green}{\underline{green}}.}
    \vspace{-1em}
    \label{tab1}
\end{table}
\noindent\textbf{Qualitative Evaluation.}
The editing results are presented in Fig.~\ref{fig: teaser},  Fig.~\ref{fig: main_experiment}, and Fig.~\ref{fig: Comparison}. Our method demonstrates precise video editing capabilities by exploring the powerful temporal information generation of the T2V model~\cite{guo2023animatediff}, achieving superior motion smoothness and enhanced realism.
\wyk{For example, the breathing of animals and the swaying leaves blown by the wind in~\cref{fig: teaser}, as well as the running person and driving cars reflecting natural sunlight in~\cref{fig: main_experiment}.}
Furthermore, our approach effectively performs shape deformation based on the editing prompt, as shown in the edited videos~(\eg, the~\textit{animals} in~\cref{fig: teaser} and the~\textit{tiger} in~\cref{fig: main_experiment}).  
\wyk{The harmony between the edited content and original video context can be observed in dynamic video demos, such as the sunlight spot on the animals in ~\cref{fig: teaser} and the reflected light on Iron Man's armor in ~\cref{fig: Comparison}.}
\begin{figure*}[t]
    \centering
    \includegraphics[width=\linewidth]{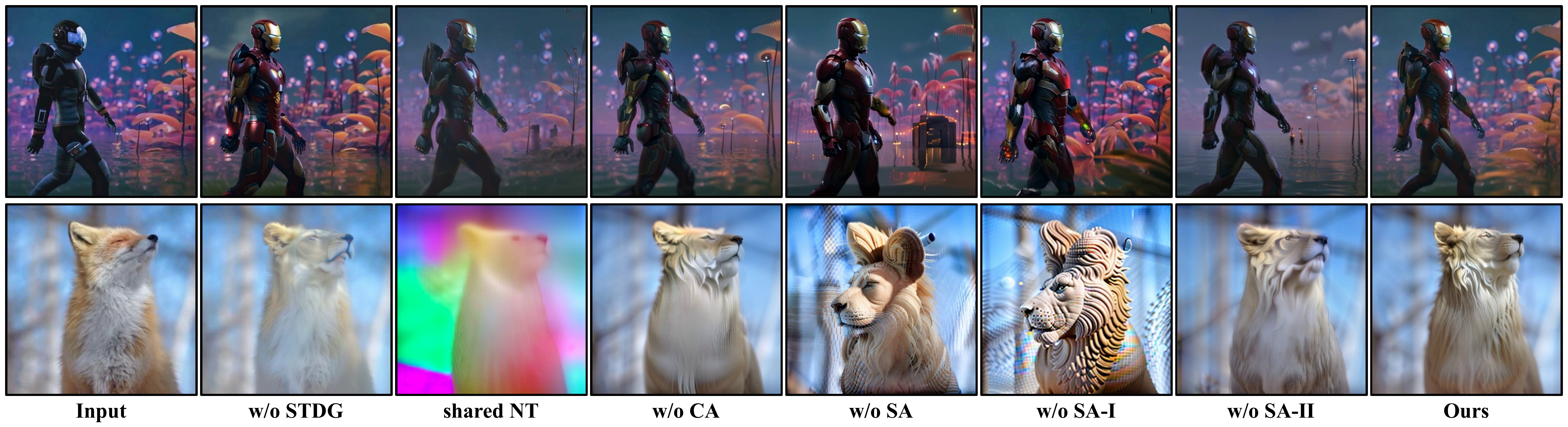} 
    \vspace{-1.5em}
    \caption{\textbf{\textit{Ablation study on editing performance.}} During editing, we use shared null-text~\textbf{(NT)} embedding, or remove ~\textbf{STDG}, the Cross Attention control module~\textbf{(CA)}, the whole Self Attention control module~\textbf{(SA)}, the Self Attention control module-\uppercase\expandafter{\romannumeral1}~\textbf{(SA-\uppercase\expandafter{\romannumeral1})}, and the Self Attention control module-\uppercase\expandafter{\romannumeral2}~\textbf{(SA-\uppercase\expandafter{\romannumeral2})} separately. 
    Our guidance and attention module can improve accuracy, fidelity, and realism. 
    }
    \vspace{-1.5em}
    \label{fig: ablation_multi_nulltext}
\end{figure*}
\begin{figure*}[t]
    \centering
    \includegraphics[width=\linewidth]{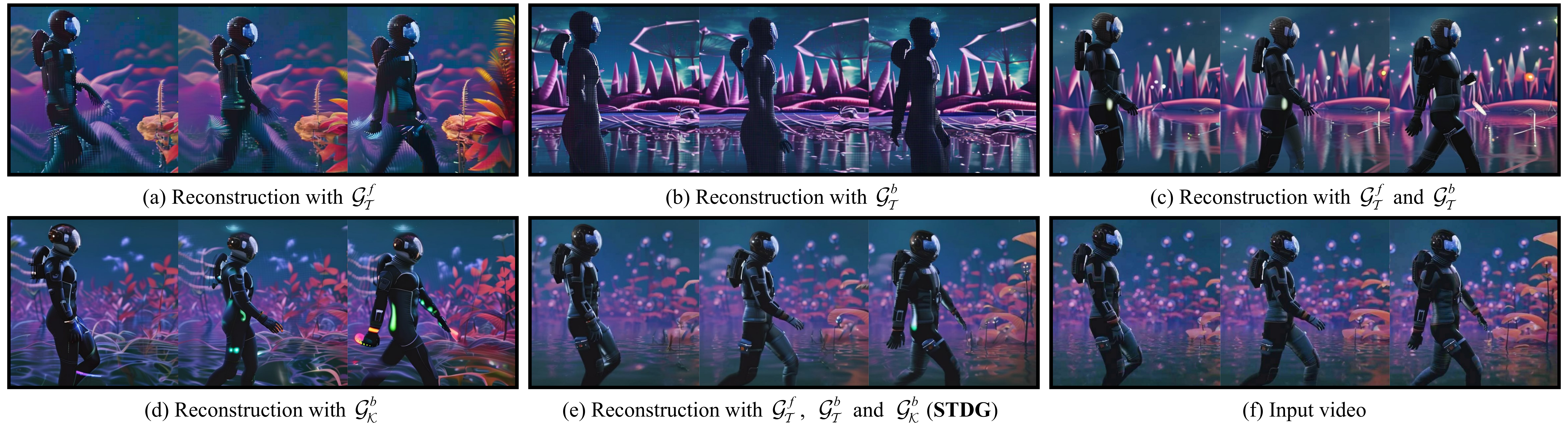} 
    \vspace{-1.5em}
    \caption{\textbf{\textit{Ablation study on STDG. }} The reconstruction performance in (c) combines the results from (a) and (b), guided by the foreground temporal guidance $\mathcal{G}_{\mathcal{T}}^{f}$ and background temporal guidance $\mathcal{G}_{\mathcal{T}}^{b}$. The performance in (e) integrates (c) and (d), incorporating the background appearance guidance $\mathcal{G}_{\mathcal{K}}^{b}$ from (d). STDG effectively guides video reconstruction, constraining the DDIM sampling trajectory.}
    \label{fig: STDG}
    \vspace{-1em}
\end{figure*}

\noindent\textbf{Quantitative Evaluation.}
We evaluate the edited videos based on four key aspects:
\textbf{Accuracy}, \textbf{Fidelity}, \textbf{Motion Smoothness}, and \textbf{Realism}.
For accuracy, we use the Pick score~\textbf{(PS)}~\cite{kirstain2023pick} to assess the alignment quality.
For fidelity, we calculate the masked PSNR (\textbf{m.P}) and LPIPS (\textbf{m.L}) to evaluate the preservation quality of the original, unedited content.
For motion smoothness~\textbf{(MS)}, we utilize VBench~\cite{huang2024vbench} to assess whether the motion in the edited video is smooth and adheres to real-world physical laws.
We also conducted a user study (\textbf{US}) to evaluate the realism of the edited videos. Nine participants were asked to rank all competing methods from best (rank \textbf{1}) to worst (rank \textbf{5}) in terms of realism and editing effectiveness, and the mean score was calculated.
As shown in Table~\ref{tab1}, our method outperforms all other methods across all metrics, demonstrating superior quantitative editing performance.

\subsection{Ablation study}
\label{sec: ablation}
\begin{table}[b]
    \vspace{-1em}
    \centering
    \small 
    \setlength{\tabcolsep}{3pt} 
    \renewcommand{\arraystretch}{0.6} 
    \scalebox{0.77}{ 
    \begin{tabular}{lccccccc}
        \toprule
        {\textbf{Abl.}} 
        &\textbf{{w/o STDG}} & \textbf{{shared NT}} & \textbf{{w/o CA}} & \textbf{{w/o SA}} & \textbf{{w/o SA-\uppercase\expandafter{\romannumeral1}}} & \textbf{{w/o SA-\uppercase\expandafter{\romannumeral2}}} & {\textbf{Ours}} \\
        \midrule
        MS~\textuparrow & 89.23\% & 97.21\% & 96.58\% & 90.37\% & 93.50\% &97.62\% & \cellcolor{lightblue}{\textbf{97.68\%}}\\
        PS~\textuparrow & 20.39 & 20.44 & 21.06 & 20.10 & 20.67 & 20.63 &\cellcolor{lightblue}{\textbf{21.64}} \\
        m.P~\textuparrow & 19.09 & 19.01& 21.13 &14.87 &16.93 &20.27 &\cellcolor{lightblue}{\textbf{21.37}}\\
        m.L~\textdownarrow &0.369 &0.346 &0.301 &0.537 &0.418 & 0.371 &\cellcolor{lightblue}{\textbf{0.270}}\\
        \bottomrule
    \end{tabular}
    }
    \vspace{-1em}
    \caption{Quantitative comparison of modules in video editing.} 
    \label{tab_metrics}
\end{table}
\noindent\textbf{Multi-frame Null Text Embedding.}
As illustrated in~\cref{fig: ablation_multi_nulltext}, multi-frame null text embeddings are crucial for editing videos with highly dynamic content (e.g., walking people or a moving fox). The incorporation of multi-frame null embeddings enhances the realism of the video and preserves more original information than shared NT, leading to significant improvements in reconstruction and editing.

\begin{figure}[t]
    \centering
    \includegraphics[width=\linewidth]{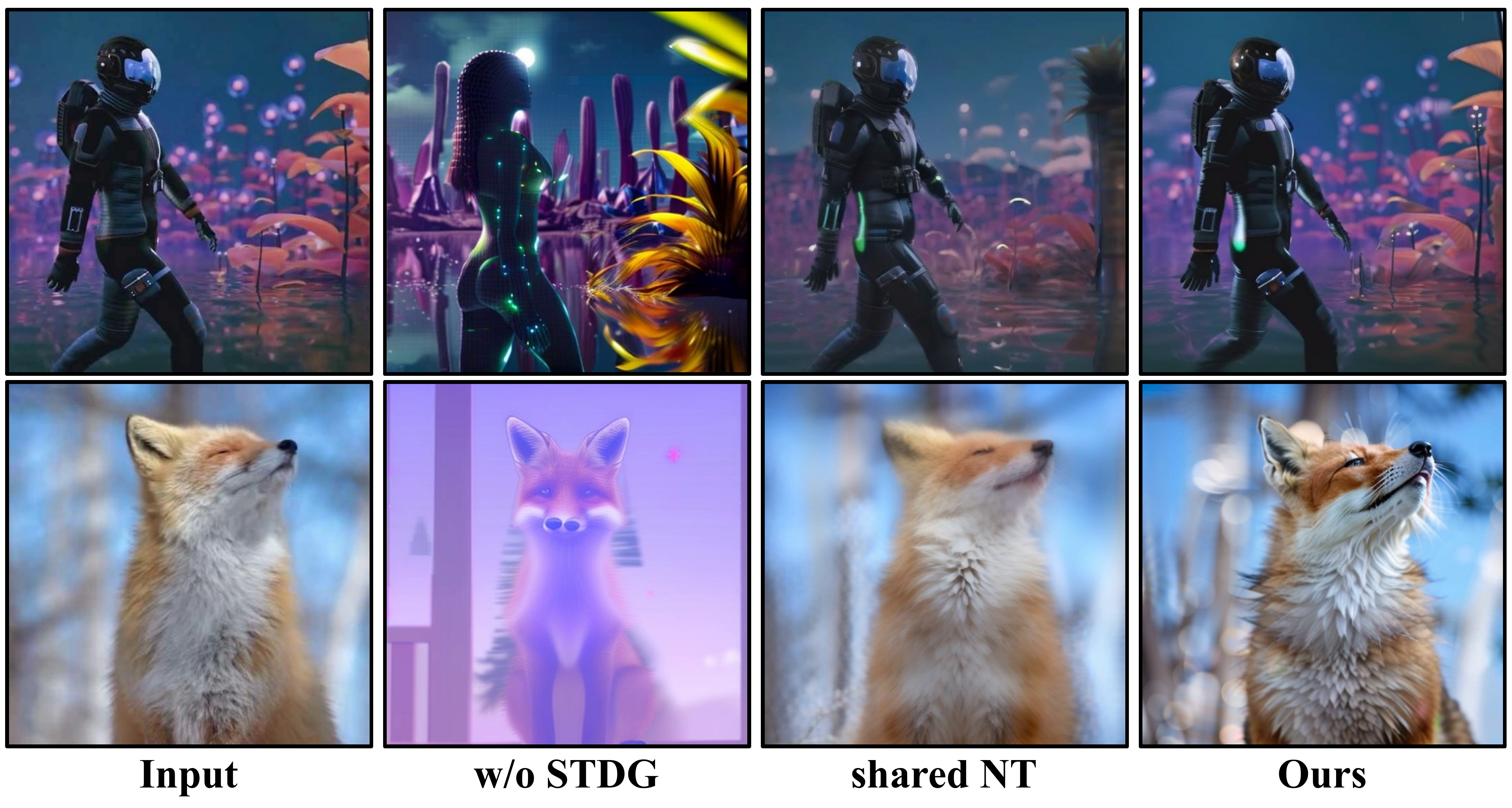} 
    \vspace{-1em}
    \caption{\textbf{\textit{Ablation study on reconstruction performance.}} We evaluate the reconstruction performance of our proposed guidance methods by either removing STDG or using shared null-text (\textbf{NT}). Our modules are crucial for effective video reconstruction.}
    \vspace{-1em}
    \label{fig: ablation_score}
\end{figure}

\noindent\textbf{Spatial-Temporal Decoupled Guidance.}
As shown in~\cref{fig: ablation_score} and ~\cref{fig: ablation_multi_nulltext}, removing the STDG significantly degrades the performance of both reconstruction and video editing. This degradation is evident from the severe color flickering and unstable video quality observed. These findings highlight the critical role of the STDG in ensuring effective video reconstruction and editing.
The quantitative results of the ablation study are shown in Tab.~\ref{tab_metrics}, 
demonstrating the effectiveness of our proposed modules.

We investigate the influence of each component of STDG in reconstructing the input video, as illustrated in ~\cref{fig: STDG}. Subfigures (a), (b), and (c) are guided by the foreground temporal guidance $\mathcal{G}_{\mathcal{T}}^{f}$, background temporal guidance $\mathcal{G}_{\mathcal{T}}^{b}$, and both respectively. When both temporal guidance components are combined, the motion reconstruction is significantly improved, as evidenced by the astronaut's hands and the lighting spots in the background. ~\cref{fig: STDG}~(d) is guided solely by the background appearance guidance $\mathcal{G}_{\mathcal{K}}^{b}$, which enhances appearance information, particularly the plants in the background. By incorporating all temporal and appearance guidance, STDG reconstructs the input video effectively, capturing both motion and appearance, as shown in ~\cref{fig: STDG}~(e).

\noindent\textbf{Attention Control Modules.}
As illustrated in ~\cref{fig: ablation_multi_nulltext}, we individually remove the attention control modules to evaluate their effectiveness in the video editing process. The results demonstrate the effectiveness of our approach in enhancing realism and fidelity. Our mutual attention strategy improves editing harmony, seamlessly integrating the edited content into the environment and context of the original video, \eg, Iron Man's armor reflecting purple light in the surroundings in ~\cref{fig: ablation_multi_nulltext}.
\vspace{-0.5em}
\section{Conclusion}
We propose VideoDirector, an approach enabling direct video editing using Text-to-Video models. Our VideoDirector develops 
spatial-temporal decoupling guidance, multi-frame null-text optimization, and an attention control strategy to harness the powerful temporal generation capability of the T2V model for precise editing. Experimental results demonstrate that our videoDirector significantly outperforms previous methods and produces results with high quality in terms of accuracy, fidelity, motion smoothness, and realism.
\section{Acknowledgements}
This work was partially supported by the National Natural Science Foundation of China (No.~U20A20185, 62372491, 62301601), the Guangdong Basic and Applied Basic Research Foundation (No.~2022B1515020103, 2023B1515120087), the Shenzhen Science and Technology Program (No.~RCYX20200714114641140), the Science and Technology Research Projects of the Education Office of Jilin Province (No.~JJKH20251951KJ).

{
    \small
    \bibliographystyle{ieeenat_fullname}
    \bibliography{main}
}



\end{document}